\documentclass[11pt]{article}
\usepackage{acl}

\usepackage{times}
\usepackage{latexsym}
\usepackage[T1]{fontenc}
\usepackage[utf8]{inputenc}
\usepackage{microtype}

\usepackage{graphicx}
\usepackage{booktabs}
\usepackage{amsmath}
\usepackage{amssymb}
\usepackage{multirow}
\usepackage{pgfplots}
\pgfplotsset{compat=1.18}
\usepgfplotslibrary{groupplots}
\usetikzlibrary{positioning,patterns}

\title{The Readout Shortcut: Positional Number Copying Dominates Arithmetic CoT Readout in Small Language Models}

\author{Ming Liu \\
  Amazon \\
  \texttt{mlliuz@amazon.com}}

\begin{document}
\maketitle

\begin{abstract}
Chain-of-thought (CoT) prompting is necessary for arithmetic in small language models, yet shuffling its steps preserves most performance. What does CoT contribute if not logical sequencing? In three 1--3B instruction-tuned LMs on GSM8K, we isolate the answer-readout stage via prefix completion and identify a positional shortcut: the model copies whichever number occupies the trailing position before the answer delimiter, regardless of intermediate reasoning. Gold-answer presence accounts for 54--92~pp of accuracy (89--92\% of each model's teacher-forcing ceiling); even on incorrect items, the final answer matches the last CoT number 95--96\% of the time. The copy channel takes precedence over retained-context completion: replacing the trailing number with a wrong value collapses accuracy to near-zero despite correct intermediates, yet removing it recovers 5--32~pp above that floor---even single-step arithmetic the model can otherwise perform is suppressed when a copyable number is present. Qwen and Llama copy novel distractors 87--95\% of the time; Gemma gates selectively. Head-level ablation implicates architecture-specific head sets; the effect replicates on GSM-Symbolic. On non-arithmetic BBH tasks, shuffle retention drops sharply; at 7--8B, content-selective gating emerges. Step-level faithfulness evaluations risk conflating positional answer transport with genuine computation---a failure mode for CoT-based oversight.
\end{abstract}

\begin{figure}[t]
\centering
\begin{tikzpicture}[
  font=\scriptsize,
  stp/.style={draw, fill=gray!15, rounded corners=1pt, minimum height=4mm, minimum width=6mm, inner sep=1pt},
  num/.style={draw, fill=blue!20, rounded corners=1pt, minimum height=4mm, minimum width=6mm, inner sep=1pt, font=\scriptsize\bfseries},
  bad/.style={draw, fill=red!25, rounded corners=1pt, minimum height=4mm, minimum width=6mm, inner sep=1pt, font=\scriptsize\bfseries},
  ans/.style={draw, thick, rounded corners=1pt, minimum height=4mm, minimum width=6mm, inner sep=1pt, font=\scriptsize\bfseries},
  delim/.style={font=\scriptsize\ttfamily},
  arr/.style={->, >=stealth, thick},
]
% Row 1: Clean CoT
\node[anchor=west] at (0,1.55) {\textbf{(a) Clean CoT}};
\node[stp] (s1) at (0.3,1.0) {$s_1$};
\node[stp, right=1mm of s1] (s2) {$s_2$};
\node[stp, right=1mm of s2] (sd) {$\cdots$};
\node[num,  right=1mm of sd] (g1) {72};
\node[delim, right=1mm of g1] (d1) {\#\#\#\#};
\node[ans, fill=blue!20, right=1mm of d1] (o1) {72};
\draw[arr, blue!70!black] (g1.south) to[bend right=35]
  node[below, midway, font=\scriptsize\itshape, yshift=-1pt] {positional copy} (o1.south);
% Row 2: Distractor
\begin{scope}[yshift=-1.8cm]
\node[anchor=west] at (0,1.55) {\textbf{(b) Wrong-number distractor}};
\node[stp] (t1) at (0.3,1.0) {$s_1$};
\node[stp, right=1mm of t1] (t2) {$s_2$};
\node[stp, right=1mm of t2] (td) {$\cdots$};
\node[num,  right=1mm of td] (g2) {72};
\node[bad,  right=1mm of g2] (x2) {999};
\node[delim, right=1mm of x2] (d2) {\#\#\#\#};
\node[ans, fill=red!25, right=1mm of d2] (o2) {999};
\draw[arr, red!70!black] (x2.south) to[bend right=35]
  node[below, midway, font=\scriptsize\itshape, yshift=-1pt] {copies wrong \#} (o2.south);
\end{scope}
\end{tikzpicture}
\caption{The \emph{answer-context-gated positional readout}: the model reads whichever number appears in answer-relevant context at the trailing position before the \texttt{\#\#\#\#} delimiter. (a)~When the correct answer is last, the readout yields the right output. (b)~In Qwen/Llama (1--3B), injecting a wrong number in answer context displaces gold and is copied 87--95\% of the time; Gemma instead shows stronger content gating ($P(\text{distractor}){=}.12$--$.19$; \S\ref{sec:distractor}).}
\label{fig:shortcut}
\end{figure}
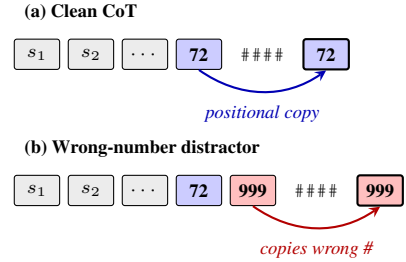

\section{Introduction}
\label{sec:intro}

Chain-of-thought (CoT) prompting \citep{wei2022chain,kojima2022large} is necessary: removing it crashes accuracy \citep{lanham2023measuring}. Yet its step order barely matters: shuffling steps retains most performance \citep{madaan2022text,wang2023towards}. This tension is well-documented but unexplained---\emph{what} in CoT drives the answer if not logical sequencing? Prior work established that CoT can be unfaithful---\citet{turpin2023language} showed biased demonstrations alter answers despite correct reasoning steps; \citet{lanham2023measuring} quantified unfaithfulness via early answering and paraphrase metrics; \citet{arcuschin2025wild} observed similar patterns in naturalistic settings---and that filler tokens partially substitute for reasoning \citep{pfau2024letsthink}; recent mechanistic studies have begun tracing information flow through CoT \citep{dutta2024how} and probing which reasoning steps matter \citep{bogdan2025anchors}. These works document \emph{that} readout can bypass reasoning but not \emph{what specific signal} the readout uses, \emph{where} in the network it operates, or \emph{whether} it actively suppresses available computation. We provide all three.

We resolve this for 1--3B instruction-tuned models on arithmetic by identifying an \emph{answer-context-gated positional readout} \citep{geirhos2020shortcut,mccoy2019hans}: the model reads whichever number appears in answer-relevant context at the trailing position before the answer delimiter, largely independent of intermediate computation. The readout requires answer-relevant framing (Appendix~\ref{sec:bare_number}) and is correctness-independent in Qwen/Llama but content-gated in Gemma (Figure~\ref{fig:shortcut}). On three architectures we show \emph{what} the model reads, \emph{why} shuffling preserves it (the answer token's positional accessibility survives when logical order does not), and \emph{where} it breaks (non-arithmetic tasks; content-selective retrieval at 7--8B). Gold-answer dependence is robust; the \emph{content gate}---operationally, the degree to which the model rejects a novel distractor number at the trailing position, measured as $1 - P(\text{distractor})$---varies from absent (Qwen: gate${\approx}0$) to strong (Gemma: gate${\approx}.85$; 7--8B models: gate${\geq}.70$; \S\ref{sec:discussion}). Concurrent behavioral work \citep{chen2026rethinking} corroborates frontier-scale shuffle-tolerance; we contribute the mechanism, which requires open-weight access.

This characterization emerges from four converging lines of evidence on three architectures (Qwen2.5-1.5B-Instruct, Llama-3.2-1B-Instruct, Gemma-2-2B-it) on GSM8K \citep{cobbe2021gsm8k}; Gemma participates fully in the corruption decomposition, shuffle hierarchy, and head-level ablation, but its position/distractor results use a teacher-forcing-passing subset (Appendix~\ref{sec:tf_fidelity}): (i)~a corruption decomposition isolates gold-answer \emph{presence} as the dominant factor (54--92~pp raw; 89--92\% ceiling-corrected), and a seven-condition causal ladder shows the copy channel takes precedence over available retained-context completion (\S\ref{sec:copy_channel}); (ii)~the readout selects the trailing answer-context number with a sharp final-position jump ($+20$--$31$~pp) and novelty-permissive copying in Qwen/Llama (.87--.95 for novel numbers) vs.\ selective gating in Gemma (.12--.19; \S\ref{sec:positional_selection}); (iii)~head-level ablations reveal architecture-specific profiles from localized (Llama) to distributed (Qwen; \S\ref{sec:sensitivity}); (iv)~the shortcut is present in base weights, replicates on GSM-Symbolic, shifts toward selectivity at 7--8B, and collapses on non-arithmetic BBH (\S\ref{sec:generalization}).

\paragraph{Scope and faithfulness.} CoT use involves \emph{rationale generation} (which may compute) and \emph{answer readout} (which maps a completed prefix to the final token); our prefix-completion interventions isolate the second stage. Following \citet{lanham2023measuring}, we treat a readout as \emph{faithful} when the answer depends on intermediate computation ($\Delta_{\text{off-copy}}$) rather than surface features ($\Delta_{\text{copy}}$). We make no claim about generation-time computation---only that the readout does not faithfully use it when a trailing answer-context number is available. This undermines the premise of step-level CoT monitors \citep{lightman2023lets,chen2025reasoning,korbak2025monitorability}: the readout takes precedence over available retained-context completion.

\section{Experimental Setup}
\label{sec:setup}

\paragraph{Models.} Qwen2.5-1.5B-Instruct \citep{qwen2.5} (28L, 1536d, 12/2 GQA heads), Llama-3.2-1B-Instruct \citep{llama3} (16L, 2048d, 32/8 GQA heads), and Gemma-2-2B-it \citep{gemma2} (26L, 2304d, 8/4 GQA heads). All use grouped-query attention \citep[GQA;][]{ainslie2023gqa} with per-head $\mathbf{o}_{\text{proj}}$ (attention output projection) columns, enabling per-head ablation. The three models span distinct tokenizer families (BPE variants for Qwen/Llama; SentencePiece unigram for Gemma), so convergent findings are not attributable to shared sub-word segmentation of numbers. Gemma appears in core experiments (corruption decomposition, shuffle hierarchy, head-level ablation, base-model probe); its position/distractor results (\S\ref{sec:positional_selection}) show reduced teacher-forcing fidelity under prefix-structure perturbation (${\sim}60\%$ vs.\ ${\sim}99\%$ on the corruption pipeline; Appendix~\ref{sec:tf_fidelity}).

\paragraph{Task and protocol.} GSM8K test set \citep{cobbe2021gsm8k}, first 500 problems. Baseline greedy CoT accuracy: Qwen 67.0\%, Llama 45.6\%, Gemma 66.2\%. All experiments use teacher-forced prefix injection: a (possibly modified) CoT is injected as the beginning of the assistant turn using native chat templates, ``\texttt{\#\#\#\# }'' is appended, and the model generates the answer greedily. This is the standard protocol for causal CoT analysis, isolating the readout from generation-time confounds \citep{lanham2023measuring}.\footnote{Gemma-2-2B-it lacks native system-prompt support; we prepend the system message to the user turn.}

\paragraph{Statistical methodology.} Wilson 95\% CIs \citep{wilson1927probable} for proportions; McNemar's exact test \citep{mcnemar1947note} for paired contrasts; Holm--Bonferroni correction \citep{holm1979simple} within pre-declared confirmatory sub-families (Appendix~\ref{sec:stats_appendix}). Auxiliary analyses (induction overlap, mean-ablation, activation patching) are exploratory with uncorrected $p$-values.

\paragraph{Notation.} $\Delta_{\text{copy}}{=}P_B{-}P_A$, $\Delta_{\text{off-copy}}{=}P_C{-}P_B$, and $P(\text{residual}){=}P_A$ are \emph{accuracy contributions} (differences between conditional accuracies) that decompose $P_C$ additively; we use $\Delta$ for these counterfactual contrasts and reserve $P(\cdot)$ for event proportions ($P(\text{distractor})$, $P(\text{gold})$) measured on a single output distribution. Causal-ladder conditions D\textsubscript{rep}/D\textsubscript{trunc}/D\textsubscript{blank} are defined in \S\ref{sec:suppression}.

\section{What Drives CoT Readout?}
\label{sec:copy_channel}

Among correctly-solved items, gold-answer presence accounts for 54--92~pp (89--92\% of each model's TF ceiling; \S\ref{sec:corruption}); a causal ladder (\S\ref{sec:suppression}) reveals that correct intermediates carry a latent signal (4--29~pp above no-CoT) that is masked when a trailing number is available. We condition on baseline-correct items; the free-generation diagnostic (\S\ref{sec:off_ceiling}) shows even incorrect answers match the last CoT number at .905--.957 across all three architectures.

\subsection{Isolating the Gold-Presence Effect}
\label{sec:corruption}

Starting from correctly-solved problems (Qwen: $n{=}335$; Llama: $n{=}228$; Gemma: $n{=}331$), we construct three conditions by selectively corrupting numbers in the CoT prefix: \textbf{A (corrupt-all)} replaces all numbers including gold occurrences; \textbf{B (preserve-gold)} corrupts intermediates but leaves gold occurrences intact; \textbf{C (clean)} presents the original CoT. Corruption uses deterministic per-example seeding.\footnote{Items where corruption preserves the gold value (mostly single-digit golds in Gemma) are excluded from Condition A (Qwen/Llama: 5 each; Gemma: 51); including them yields $P_B - P_A \geq .789$ in all cases. Conditions B and C use the full sample.}

\begin{table}[t]
\centering
\small
\resizebox{\columnwidth}{!}{%
\begin{tabular}{lccc}
\toprule
\textbf{Condition / Component} & \textbf{Qwen} & \textbf{Llama} & \textbf{Gemma} \\
\midrule
A: No gold in CoT ($n_A$) & .079 [.054, .113] & .117 [.081, .165] & .037 [.021, .063] \\
B: Gold present & 1.00\phantom{0} [.989, 1.00] & .991 [.969, .998] & .580 [.526, .632] \\
C: Everything correct & 1.00\phantom{0} [.989, 1.00] & .996 [.976, .999] & .604 [.551, .655] \\
\midrule
$\Delta_{\text{copy}} = P_B - P_A$ & .921 [.889, .946] & .875 [.828, .914] & .543 [.489, .597] \\
$\Delta_{\text{off-copy}} = P_C - P_B$ & .000 [.000, .011] & .004 [.001, .024] & .024 [.009, .050] \\
$P(\text{residual}) = P_A$ & .079 [.054, .113] & .117 [.081, .165] & .037 [.021, .063] \\
\midrule
$\Delta_{\text{copy}} / P_C$ (ceiling-norm.) & .924 & .879 & .899 \\
\bottomrule
\end{tabular}}
\caption{Corruption decomposition (Qwen $n{=}335$, Llama $n{=}228$, Gemma $n{=}331$). Wilson 95\% CIs for condition accuracies (rows A--C); paired bootstrap 95\% CIs over items for derived differences ($\Delta_{\text{copy}}$, $\Delta_{\text{off-copy}}$); $P(\text{residual}){=}P_A$ uses the Condition~A interval directly. Gold-presence dominates (54--92~pp raw; 88--92\% ceiling-normalized). Gemma's lower raw $\Delta_{\text{copy}}$ reflects reduced teacher-forcing fidelity ($P_C{=}.604$) under the standard prefix format, not a weaker copy mechanism: ceiling-normalized copy strength converges across architectures (bottom row). $\Delta$ denotes accuracy \emph{attributable to} each component (i.e., a difference between conditional accuracies), not an event probability.}
\label{tab:decomposition}
\end{table}

Gold-presence ($\Delta_{\text{copy}} = P_B - P_A$) accounts for 54--92~pp raw across three architectures (Table~\ref{tab:decomposition}). Gemma's lower raw value reflects its reduced TF-fidelity ($P_C{=}.604$; the standard prefix format interacts with Gemma's chat template). Ceiling-normalized copy strength ($\Delta_{\text{copy}}/P_C$) converges at 88--92\%, indicating a quantitatively comparable mechanism operating near each model's achievable ceiling. Intermediate-step computation ($\Delta_{\text{off-copy}} = P_C - P_B$) is at noise floor for Qwen/Llama and modest for Gemma. Conditions A and B differ only in gold-answer presence, cleanly isolating this effect. Whether the copy is positional vs.\ content-selective is tested in \S\ref{sec:distractor}.

\subsection{Off-Ceiling Stress Test}
\label{sec:off_ceiling}

To create headroom for measuring $\Delta_{\text{off-copy}}$, we swap in a weaker configuration---Qwen-base, 0-shot ($P_B{=}0.735$, $n{=}200$)---with 26~pp of headroom, yet $\Delta_{\text{off-copy}}{=}0.020$ (bootstrap 95\% CI $[-.015, .060]$). This is an order of magnitude below the $.730$ copy contribution. Correct intermediate steps add essentially no information to the readout.

If trailing-number alignment were a prefix-injection artifact, it should disappear under unconstrained decoding. It does not (Table~\ref{tab:free_gen_main}): under free generation ($n{=}500$), all three models' final answers match the last CoT number 96--97\% of the time. Critically, on items the model gets \emph{wrong}, the match rate is $.905$--$.957$---under a ``compute-then-write'' alternative, errors would distribute across non-trailing numbers, but they are pinned to the trailing position. Conditioning on whether gold occupies the last CoT position reveals a near-deterministic gate: accuracy is $.991$--$.997$ when gold is last vs.\ $.004$--$.030$ when it is not. This full-distribution analysis confirms the shortcut operates identically on items the model gets wrong, independent of any teacher-forcing intervention.

\begin{table}[t]
\centering
\small
\begin{tabular}{lccc}
\toprule
\textbf{Metric} & \textbf{Qwen} & \textbf{Llama} & \textbf{Gemma} \\
\midrule
Baseline accuracy & .672 & .448 & .662 \\
Final answer $=$ last CoT num & .972 & .974 & .960 \\
Acc $\mid$ gold is last & .997 & .991 & .994 \\
Acc $\mid$ gold is \emph{not} last & .030 & .004 & .023 \\
Ans $=$ last $\mid$ incorrect & .945 & .957 & .905 \\
\bottomrule
\end{tabular}
\caption{Free-generation analysis ($n{=}500$, unconstrained greedy decoding, no prefix injection). Accuracy is near-perfectly predicted by whether gold occupies the last CoT position. Even on incorrect items, the answer reflexively matches the trailing number---the same positional-copy signature confirmed on all three architectures.}
\label{tab:free_gen_main}
\end{table}

\subsection{Copy Masks Retained-Context Computation}
\label{sec:suppression}

The decomposition above leaves a residual question: does $\Delta_{\text{off-copy}} \approx 0$ reflect genuine inability to complete from retained context, or does the copy channel \emph{override} an available completion pathway? We disambiguate with four additional conditions that vary the trailing slot independently of intermediate-step correctness.

\paragraph{D-replace: correct intermediates, wrong trailing number.} We hold intermediates correct (as in Condition~B) but replace gold-answer occurrences with a wrong number, yielding Condition~\textbf{D\textsubscript{rep}}. Accuracy collapses to $.076$ (Qwen, $n{=}328$), $.084$ (Llama, $n{=}214$), $.010$ (Gemma, $n{=}312$)---within ${\leq}1.1$~pp of Condition~A ($.073$, $.079$, $.000$), where intermediates are also corrupted (Table~\ref{tab:causal_ladder}; pairwise McNemar $p > .01$; all nine \S\ref{sec:suppression} confirmatory contrasts yield raw $p < 10^{-8}$ and remain significant under Holm--Bonferroni correction at any family size up to the full 23-contrast set). The wrong number is copied with $P(\text{distractor}) = .924$ (Qwen), $.911$ (Llama), $.574$ (Gemma). Clean intermediates provide no benefit when a trailing distractor is present.

\paragraph{D-truncate: intermediates only, no trailing number.} To probe whether this completion ability exists but is masked, \textbf{D\textsubscript{trunc}} truncates the CoT before the first gold-answer occurrence (at the preceding sentence boundary, retaining ${\sim}80\%$ of tokens), leaving only correct intermediate steps with no trailing number. Accuracy rises to $.399$ (Qwen, $n{=}308$), $.183$ (Llama, $n{=}208$), $.058$ (Gemma, $n{=}292$)---a $+32$/$+10$/$+5$~pp gain over D\textsubscript{rep}: the same intermediates yield non-trivial accuracy once the trailing distractor is removed.\footnote{Only 1--2\% of D\textsubscript{trunc} items have their new trailing number equal to gold (Qwen 5/308, Llama 4/208, Gemma 3/292); filtered rates (.389/.172/.055) are virtually identical to overall rates.} A depth partition (Table~\ref{tab:dtrunc_depth}) shows 74--77\% of recovered items are 1-op reachable from the last visible intermediate; the shortcut thus suppresses even single-step arithmetic the model can otherwise perform.

\paragraph{Controls: No-CoT and D-blank.} Two controls close remaining attack surfaces. \textbf{No-CoT} (direct answer, no rationale prefix) yields $.106$ (Qwen), $.019$ (Llama), $.014$ (Gemma). The $29$~pp gap D\textsubscript{trunc}${-}$No-CoT on Qwen (McNemar $p < 10^{-12}$) shows the truncated intermediates carry genuine signal---the model is not re-solving from the question alone. \textbf{D\textsubscript{blank}} appends a content-free closing sentence to the truncated prefix (restoring CoT format without a trailing number): $.461$, $.197$, $.213$. D\textsubscript{blank}${\approx}$D\textsubscript{trunc} on Qwen and Llama confirms that format completeness is not the confound; removing the trailing number, not the closing template, releases the available completion pathway.

\begin{table}[t]
\centering
\small
\setlength{\tabcolsep}{3pt}
\renewcommand{\arraystretch}{1.05}
\resizebox{\columnwidth}{!}{%
\begin{tabular}{l|ccc|cc|cc}
\toprule
& \multicolumn{3}{c|}{\textit{Floor}} & \multicolumn{2}{c|}{\textit{Recompute}} & \multicolumn{2}{c}{\textit{Ceiling}} \\
\cmidrule(lr){2-4}\cmidrule(lr){5-6}\cmidrule(lr){7-8}
\textbf{Model} & \textbf{No-CoT} & \textbf{A} & \textbf{D\textsubscript{rep}} & \textbf{D\textsubscript{trunc}} & \textbf{D\textsubscript{blank}} & \textbf{B} & \textbf{C} \\
\midrule
Qwen-1.5B  & .106 & .073 & .076 & .399 & .461 & 1.000 & 1.000 \\
Llama-1B   & .019 & .079 & .084 & .183 & .197 & .991 & .991 \\
Gemma-2B   & .014 & .000 & .010 & .058 & .213 & .571 & .590 \\
\midrule
\multicolumn{8}{l}{\footnotesize \emph{Copy-override gap} (D\textsubscript{trunc}${-}$D\textsubscript{rep}): Qwen \textbf{+32}pp, Llama \textbf{+10}pp, Gemma \textbf{+5}pp} \\
\multicolumn{8}{l}{\footnotesize \emph{Retained-context contribution} (D\textsubscript{trunc}${-}$No-CoT): Qwen \textbf{+29}pp, Llama \textbf{+16}pp, Gemma \textbf{+4}pp} \\
\multicolumn{8}{l}{\footnotesize $P(\text{dist})$ in D\textsubscript{rep}: Qwen .924, Llama .911, Gemma .574} \\
\bottomrule
\end{tabular}}
\caption{Causal ladder ($P(\text{gold})$, baseline-correct, per-experiment common-index subsets). Three regimes: \textbf{floor} (No-CoT, A, D\textsubscript{rep})---trailing number absent or wrong, accuracy collapses; \textbf{recompute} (D\textsubscript{trunc}, D\textsubscript{blank})---removing distractor reveals available completion pathway; \textbf{ceiling} (B, C)---gold presence. Gemma B/C lower than Table~\ref{tab:decomposition} because truncation-length and leak filters disproportionately exclude Gemma items; per-condition $n$ in Appendix~\ref{sec:stats_appendix}.}
\label{tab:causal_ladder}
\end{table}

\paragraph{Copy-over-recompute.} Together: (a)~retained-context completion exists (D\textsubscript{trunc} $\gg$ No-CoT, Qwen $+29$~pp, $p{<}10^{-12}$); (b)~the copy channel takes precedence (D\textsubscript{rep} $\approx$ A despite identical intermediates); (c)~the precedence is answer-context-gated (D\textsubscript{blank} $\approx$ D\textsubscript{trunc}; cf.\ \S\ref{sec:end_position}, Appendices~\ref{sec:bare_number},~\ref{sec:novel_delimiter}). ``Override'' is operational: the output matches the trailing number rather than the completion pathway's prediction; we remain agnostic between active suppression and passive monopolization. The magnitude tracks copy strength: Qwen ($P(\text{distractor}){=}.924$) shows the largest D\textsubscript{trunc} gain ($+32$~pp); Gemma ($.574$) the smallest, with a $+15$~pp D\textsubscript{blank}${-}$D\textsubscript{trunc} gap suggesting additional format sensitivity. For CoT-based oversight, this is the load-bearing finding: evaluations that vary intermediates while preserving the trailing answer slot systematically underestimate available computation.

\section{How Is the Answer Selected?}
\label{sec:positional_selection}

Three converging tests distinguish end-anchored copy from content-selective retrieval or generic recency \citep[cf.][]{liu2024lost}: both models copy wrong numbers placed at the end (\S\ref{sec:distractor}); fixing position recovers the shuffle drop with a sharp final-slot jump (\S\ref{sec:end_position}); and a seven-level hierarchy shows performance tracks token accessibility (\S\ref{sec:hierarchy}).

\subsection{The Model Copies Wrong Numbers}
\label{sec:distractor}

If the readout is content-selective for gold, it should resist distractors \citep[cf.][]{shi2023distracted}. Starting from the keep-end condition (accuracy ${\approx}1.0$), we append a wrong number between the answer step and the delimiter. We pre-register a decision rule: $P(\text{distractor}) > 0.70$ rejects gold-specific retrieval as the dominant readout (a content-selective readout predicts $P(\text{distractor})$ near~0; $.70$ ensures rejection requires substantial copy-dominance; observed values exceed this by 17--25~pp, so conclusions are insensitive to the exact threshold within $[.50, .85]$).\footnote{Gemma's Experiments~3--4 are excluded from headline comparisons; teacher-forcing fidelity falls below the $.80$ threshold (Appendix~\ref{sec:tf_fidelity}). On the fidelity-passing subset ($n{=}196$), $P(\text{distractor}){=}.12$--$.19$.}

Five conditions test adjacent (gold$\pm$1), random (same digit-count), and control distractors:

\begin{table}[t]
\centering
\small
\resizebox{\columnwidth}{!}{%
\begin{tabular}{lcccc}
\toprule
 & \multicolumn{2}{c}{\textbf{Qwen} ($n{=}335$)} & \multicolumn{2}{c}{\textbf{Llama} ($n{=}227$)} \\
\cmidrule(lr){2-3} \cmidrule(lr){4-5}
\textbf{Cond.} & $P(\text{gold})$ & $P(\text{dist})$ & $P(\text{gold})$ & $P(\text{dist})$ \\
\midrule
C0 (baseline) & 1.00 & --- & .982 & --- \\
C0b (filler) & .782 & --- & .978 & --- \\
C1 (adjacent) & .051 & \textbf{.949} [.920,.968] & .053 & \textbf{.947} [.910,.970] \\
C2 (random) & .119 & \textbf{.881} [.841,.911] & .123 & \textbf{.872} [.823,.910] \\
C3 (gold dup.) & 1.00 & --- & .996 & --- \\
\bottomrule
\end{tabular}}
\caption{Distractor injection results. C0: baseline (no distractor); C0b: non-numeric filler; C1: adjacent wrong number (gold${\pm}1$); C2: random wrong number; C3: gold duplicated (positive control). Both models copy a novel wrong number 87--95\% of the time.}
\label{tab:distractor}
\end{table}

Both novel-distractor conditions exceed the $.70$ threshold by wide margins (Table~\ref{tab:distractor}; one-sided exact binomial: C1 $p < 10^{-15}$; C2 $p < 10^{-8}$). With novel numbers, the mechanism is predominantly not content-selective for gold in the 1--3B regime. A ${\sim}7$~pp C1--C2 gap suggests minor content sensitivity, well below $.70$. The asymmetric C0b sensitivity (Qwen $-22$~pp vs.\ Llama $-0.4$~pp from non-numeric filler) mirrors the topology contrast in \S\ref{sec:sensitivity}: Qwen's distributed copy circuit is destabilized by any trailing material, whereas Llama's concentrated circuit implements a tighter numeric-target selector.

Novel-delimiter controls ($P(\text{distractor}) \geq .90$; Appendix~\ref{sec:novel_delimiter}) rule out delimiter surface form. The ``answer-context-gated'' qualifier is established by a framing-dissociation experiment (Table~\ref{tab:bare_number_main}): bare numbers at the trailing position yield low copying (.45/.26), while answer-relevant inline framing (``actually it should be X'') recovers high copying (.77/.85)---same number, same position, but $P(\text{distractor})$ swings ${\sim}2{\times}$ based purely on whether the surrounding text is answer-relevant. The readout is gated by answer-context framing, not raw position or template surface form.

\begin{table}[t]
\centering
\small
\resizebox{\columnwidth}{!}{%
\begin{tabular}{lcccc}
\toprule
& \multicolumn{2}{c}{\textbf{Qwen}} & \multicolumn{2}{c}{\textbf{Llama}} \\
\cmidrule(lr){2-3} \cmidrule(lr){4-5}
\textbf{Framing condition} & $P(\text{dist})$ & $P(\text{gold})$ & $P(\text{dist})$ & $P(\text{gold})$ \\
\midrule
F1: Template (``the answer is X'') & .893 & .107 & .877 & .123 \\
F2: Bare number only & .448 & .549 & .260 & .709 \\
F3: Non-answer (``Note: X'') & .206 & .788 & .339 & .652 \\
F4: Inline (``should be X'') & .770 & .110 & .846 & .137 \\
\bottomrule
\end{tabular}}
\caption{Framing-dissociation control ($n{=}335$ Qwen; $n{=}227$ Llama). F-codes denote framing conditions (distinct from distractor C-codes in Table~\ref{tab:distractor}). The readout is \emph{answer-context-gated}: bare trailing numbers (F2) and non-answer framing (F3) fall well below the $.70$ threshold; answer-relevant inline text (F4) triggers copying comparably to standard templates (F1). This dissociates the mechanism from both raw positional recency and template-specific parsing.}
\label{tab:bare_number_main}
\end{table}

\paragraph{Intermediate-result distractor.} When the trailing number is an \emph{intermediate computation result} (e.g., a subtotal from an earlier step)---already bound to a semantic role in the CoT---the models diverge ($n{=}200$). Qwen copies it 86\% [.81,.90], consistent with novelty-permissive copying. Llama copies it only 25\% [.20,.31], recovering the gold answer 74.5\% of the time---evidence of a weak repetition-suppression filter layered on the dominant positional mechanism (25\% remains far above the ${\sim}1\%$ no-distractor error floor, confirming positional copying still dominates). Architectures thus differ not in \emph{whether} the shortcut is answer-context-gated but in \emph{how sharply} the gate discriminates non-novel numerals: Qwen~$<$~Llama~$<$~Gemma.

\subsection{Fixing Position Recovers the Shuffle Drop}
\label{sec:end_position}

If the readout is end-anchored, most of the shuffle drop should be recoverable by controlling where the answer step appears. We test four conditions: \textbf{ordered}, \textbf{full\_shuffle}, \textbf{keep\_end} (non-answer steps shuffled, answer fixed at end), and \textbf{move\_front} (answer at front).

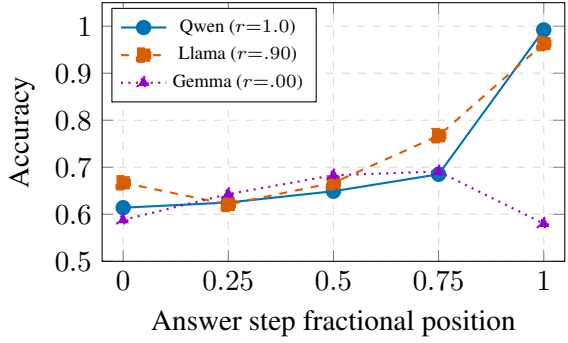
\begin{figure}[t]
\centering
\begin{tikzpicture}
\begin{axis}[
    width=\columnwidth,
    height=5cm,
    xlabel={Answer step fractional position},
    ylabel={Accuracy},
    xmin=-0.05, xmax=1.05,
    ymin=0.5, ymax=1.05,
    xtick={0,0.25,0.5,0.75,1.0},
    ytick={0.5,0.6,0.7,0.8,0.9,1.0},
    legend style={at={(0.02,0.98)}, anchor=north west, font=\scriptsize},
    grid=major,
    grid style={dashed, gray!30},
    mark size=2.5pt,
]
\addplot[color={rgb,1:red,0.0;green,0.45;blue,0.70}, thick, solid, mark=*] coordinates {(0,.614) (0.25,.625) (0.5,.649) (0.75,.685) (1.0,.992)};
\addplot[color={rgb,1:red,0.85;green,0.37;blue,0.01}, thick, dashed, mark=square*] coordinates {(0,.667) (0.25,.621) (0.5,.667) (0.75,.767) (1.0,.963)};
\addplot[color={rgb,1:red,0.58;green,0.0;blue,0.83}, thick, dotted, mark=triangle*] coordinates {(0,.588) (0.25,.643) (0.5,.683) (0.75,.691) (1.0,.580)};
\legend{Qwen ($r{=}1.0$), Llama ($r{=}.90$), Gemma ($r{=}.00$)}
\end{axis}
\end{tikzpicture}
\caption{Answer-position curve (5-position sweep, $n{=}123$--$315$). Qwen/Llama show a shallow-gradient-then-jump profile: accuracy rises non-monotonically over 0--75\% then jumps $+31$/$+20$ pp at position 1.0 (end). Gemma is flat (teacher-forcing fidelity issue; Appendix~\ref{sec:tf_fidelity}). Discrete conditions: keep\_end (pos$=$1.0) recovers ordered accuracy within ${\leq}1$~pp (McNemar $p \geq .25$).}
\label{fig:position}
\end{figure}

Keeping the answer step at the end recovers 99\% (Qwen) / 93\% (Llama) of the shuffle drop (Figure~\ref{fig:position}); ordered and keep\_end differ by ${\leq}1$~pp (McNemar $p \geq .25$). The shallow-gradient-then-jump profile---$+7$/$+10$~pp over positions 0--75\%, then $+31$/$+20$~pp at position 1.0---is inconsistent with smooth recency weighting.

A positional-encoding control (Appendix~\ref{sec:positional_control}) rules out RoPE \citep{su2024roformer} artifacts: monotonic position-ID stretching ($2.5{\times}$) causes no measurable accuracy loss ($n{=}66$), showing that position-encoding perturbation does not account for the shuffle drop.

\subsection{Performance Tracks Token Accessibility}
\label{sec:hierarchy}

The binary keep\_end/move\_front contrast is coarse. A seven-level shuffle hierarchy decomposes the perturbation along structural granularity: step-level (step\_shuffle, reverse\_order), word-level (within\_step, word\_shuffle), and token-level (token\_shuffle), plus ordered and no\_cot anchors.

\begin{figure}[t]
\centering
\begin{tikzpicture}
\begin{axis}[
    width=\columnwidth,
    height=5.2cm,
    ybar,
    bar width=3.5pt,
    ylabel={Retention (\%)},
    symbolic x coords={Ordered,Within-step,Step-shuffle,Word-shuffle,Reverse,Token-shuffle,No-CoT},
    xtick=data,
    x tick label style={rotate=30, anchor=east, font=\scriptsize},
    ymin=-10, ymax=110,
    ytick={0,25,50,75,100},
    extra y ticks={0},
    extra y tick style={grid=major, grid style={solid, gray!70, line width=0.6pt}, tick label style={opacity=0}},
    legend style={at={(0.98,0.98)}, anchor=north east, font=\scriptsize, legend columns=3},
    enlarge x limits=0.08,
    grid=major,
    grid style={dashed, gray!30},
    every axis plot/.append style={fill opacity=0.85},
]
\addplot[fill={rgb,1:red,0.0;green,0.45;blue,0.70}, draw={rgb,1:red,0.0;green,0.30;blue,0.55}, postaction={pattern=north east lines, pattern color=white},
    error bars/.cd, y dir=both, y explicit, error bar style={line width=0.5pt, gray!70!black}, error mark options={rotate=90, mark size=2pt, line width=0.4pt}]
    coordinates {(Ordered,100) +- (0,0) (Within-step,93) +- (1.9,2.2) (Step-shuffle,68) +- (3.6,3.0) (Word-shuffle,44) +- (4.2,3.9) (Reverse,14) +- (4.6,5.2) (Token-shuffle,-4) +- (1.8,1.6) (No-CoT,0) +- (3.6,2.9)};
\addplot[fill={rgb,1:red,0.85;green,0.37;blue,0.01}, draw={rgb,1:red,0.65;green,0.27;blue,0.0}, postaction={pattern=crosshatch, pattern color=white},
    error bars/.cd, y dir=both, y explicit, error bar style={line width=0.5pt, gray!70!black}, error mark options={rotate=90, mark size=2pt, line width=0.4pt}]
    coordinates {(Ordered,100) +- (0.5,0.9) (Within-step,74) +- (3.8,3.9) (Step-shuffle,83) +- (3.1,3.7) (Word-shuffle,38) +- (4.6,4.4) (Reverse,60) +- (7.0,6.6) (Token-shuffle,16) +- (3.5,2.8) (No-CoT,0) +- (2.3,1.8)};
\addplot[fill={rgb,1:red,0.58;green,0.0;blue,0.83}, draw={rgb,1:red,0.40;green,0.0;blue,0.60}, postaction={pattern=dots, pattern color=white},
    error bars/.cd, y dir=both, y explicit, error bar style={line width=0.5pt, gray!70!black}, error mark options={rotate=90, mark size=2pt, line width=0.4pt}]
    coordinates {(Ordered,100) +- (2.6,3.0) (Within-step,48) +- (5.1,3.9) (Step-shuffle,87) +- (2.4,3.1) (Word-shuffle,24) +- (3.4,4.1) (Reverse,41) +- (5.7,5.8) (Token-shuffle,1) +- (0.5,1.3) (No-CoT,0) +- (1.3,1.0)};
\legend{Qwen, Llama, Gemma}
\end{axis}
\end{tikzpicture}
\caption{Shuffle hierarchy with bootstrap 95\% CIs ($n{=}228$--$335$ items). Retention (no-CoT-anchored) $= (P_\text{cond} - P_\text{no\_cot})/(P_\text{ord} - P_\text{no\_cot})$ anchors 0\% at no-CoT and 100\% at ordered. Step-level conditions preserve the answer token and retain 68--87\%; token-level shuffling destroys token identity and collapses to no-CoT. Negative values indicate active interference.}
\label{fig:hierarchy}
\end{figure}
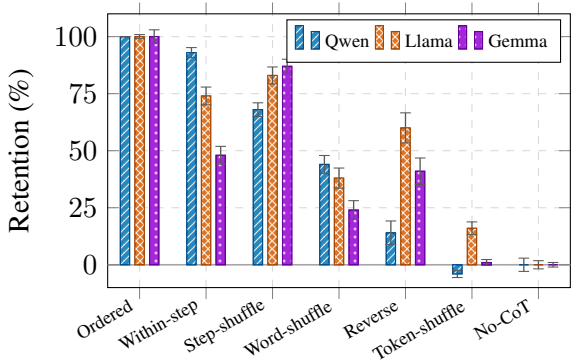

The hierarchy is monotone in answer-token accessibility (Figure~\ref{fig:hierarchy}): within each granularity level, conditions preserving the answer at the end (within\_step) outperform those displacing it (reverse\_order), so accessibility---not logical coherence---is the controlling variable. Self-generated CoT shuffle replicates the pattern on $N{=}2$ (Appendix~\ref{sec:selfgen_shuffle}).

\section{Architectural Variation in Head-Level Copy Sensitivity}
\label{sec:sensitivity}

Three interventions---zero-ablation, mean-ablation, and activation patching---converge on architecture-specific copy-sensitive head sets (Table~\ref{tab:ablation}). All three architectures participate in zero-ablation and mean-ablation; activation patching is reported for Qwen (the most distributed profile, where convergent evidence is most diagnostic). Gemma's zero-ablation results (K$_{50}{=}5$, L14H4 single-head $-19.4$~pp) and induction-head overlap ($p{<}0.01$) are fully included.

\subsection{Zero-Ablation Reveals Three Distinct Sensitivity Profiles}
\label{sec:ablation}

Following \citet{wu2024retrieval}, we rank heads by mean attention mass from the last prefix token to gold-answer tokens ($n{=}100$ held-out), then zero their $\mathbf{o}_{\text{proj}}$ output columns on a disjoint evaluation set ($n{=}100$), ablating at query-head granularity under GQA. Specificity is tested with random-5 head sets (Llama: 20 layer-stratified, extended to $n{=}1{,}000$ permutation; Table~\ref{tab:ablation}).

\begin{table}[t]
\centering
\small
\resizebox{\columnwidth}{!}{%
\begin{tabular}{lccc}
\toprule
\textbf{Condition} & \textbf{Qwen} & \textbf{Llama} & \textbf{Gemma} \\
\midrule
Baseline & 1.000 & .990 & .697 \\
Top-5 ablated & 1.000 ($-$0pp) & \textbf{.790} ($-$20pp) & \textbf{.648} ($-$4.9pp) \\
Random-5 control & 1.000 ($-$0pp) & .990 ($\bar\Delta{=}$0pp) & .709 ($+$1.2pp) \\
\midrule
Max single-head & $-$0pp & $-$0pp & $-$19.4pp (L14H4) \\
\bottomrule
\end{tabular}}
\caption{Zero-ablation sensitivity ($n{=}100$ Qwen/Llama; $n{=}165$ Gemma). Gemma's lower baseline (.697) reflects its TF-fidelity-passing evaluation subset. Llama: 20pp drop from 5 heads, zero from random controls (permutation $p{<}.001$, $n{=}1{,}000$). Gemma: L14H4 single-head $-19.4$pp. Qwen: 0pp from top-5; cumulative ablation collapses at $K_{50}{=}17$, and mean-ablation shows real signal ($-33.5$~pp at $K{=}20$; \S\ref{sec:mechanistic}).}
\label{tab:ablation}
\end{table}

Three profiles emerge (Table~\ref{tab:ablation}): Llama is \emph{localized} (L10--L15; permutation $p{<}.001$, $n{=}1{,}000$); Gemma is \emph{concentrated but compensable} (L14H4 dominant under zero-ablation, collectively load-bearing under mean-ablation; \S\ref{sec:mechanistic}); Qwen is \emph{distributed} ($K_{50}{=}17$, sharp phase transition at $K{\approx}14$--$15$). After ablating Llama's top-5 ($n{=}100$, acc$.{=}.79$; 21 failures), $0/21$ are verbatim copies; $14/21$ produce $2{\times}$/$3{\times}$ gold---a shift from copy-errors to recomputation-errors, indicating selectivity for the copy pathway rather than a general arithmetic deficit.

\subsection{Converging Mechanistic Evidence}
\label{sec:mechanistic}

To address OOD concerns with zero-ablation \citep{zhang2024patching}, mean-ablation (replacing each head's output with its mean over a held-out reference set) reproduces copy-specific top-$K$ drops on all three architectures: Gemma $-52$~pp at $K{=}20$, Qwen $-33.5$~pp ($K_{50}{=}11$, retaining $43\%$ of zero-ablation drop), Llama $-6.5$~pp peak. Single-head effects diverge between protocols (Gemma L14H4: $-19.4$~pp zero vs.\ $-0.5$~pp mean), but aggregate results converge. Activation patching on Qwen recovers 61\% of the shuffle gap (Appendix~\ref{sec:patching_appendix}). Content-tracking and induction-head analyses (Appendix~\ref{sec:mechanistic_appendix}) corroborate a convergent-to-diffuse gradient: Gemma/Llama show significant induction/copy overlap ($p{<}0.01$); Qwen's is near-chance. We apply \citet{wu2024retrieval}'s attention-mass ranking outside its original long-context setting; relative to prior arithmetic-circuit work \citep{nikankin2025arithmetic,dziri2023faith,stolfo2023mechanistic,mcdougall2023copy}, we document a regime where the readout operates largely independently of internal arithmetic.

\section{Generalization and Boundaries}
\label{sec:generalization}

We test memorization, instruction-tuning artifact, scale, and task specificity.

\subsection{Novel-Instantiation Replication (GSM-Symbolic)}
\label{sec:contamination}

GSM-Symbolic \citep{mirzadeh2024gsmsymbolic} (main config, test split) regenerates GSM8K-style problems from symbolic templates with numeric values drawn outside the GSM8K distribution, eliminating verbatim overlap with pre-training data. Step-shuffle retains 73\% (Qwen, $n{=}74$) / 81\% (Llama, $n{=}86$)\footnote{Qwen ceiling-corrected; Llama uncorrected (ordered acc.\ = 1.0).}---consistent with the GSM8K values (68\% / 83\%). On SVAMP \citep{patel2021nlp}, a structurally distinct arithmetic word-problem benchmark ($n{=}300$), the copy decomposition replicates: $\Delta_{\text{copy}} = 1.00$ (Qwen), $.990$ (Llama), $.603$ (Gemma; reduced by TF-fidelity; Appendix~\ref{sec:svamp}). The copy phenomenon is not an artifact of memorization or GSM8K-specific distributional features.

\subsection{Base-Model Origin}
\label{sec:base_model}

The copy mechanism is present in base-model weights (Qwen-base $\Delta_{\text{copy}}{=}.730$; Gemma-base $.335$; Llama-base $.975$ with 4-shot; Appendix~\ref{sec:base_model_appendix}). Instruction-tuning amplifies copy strength (Gemma $1.6{\times}$ raw, $.34{\to}.54$; ceiling-normalized $.77{\to}.90$) but does not create the mechanism. Together with the novel-delimiter control (Appendix~\ref{sec:novel_delimiter}) and GSM-Symbolic replication (\S\ref{sec:contamination}), these results rule out IT-introduced format prior and delimiter-specific surface-form learning.

\subsection{At 7--8B, Gold-Presence Persists but Copying Becomes Selective}
\label{sec:scale}

Scaling up by $5{\times}$--$8{\times}$, we test Qwen2.5-7B-Instruct and Llama-3.1-8B-Instruct.\footnote{Gemma-2-9B-it excluded: pipeline debugging incomplete within submission timelines (Appendix~\ref{sec:tf_fidelity}); not evidence about Gemma scaling.}

\begin{table}[t]
\centering
\small
\resizebox{\columnwidth}{!}{%
\begin{tabular}{lccc}
\toprule
\textbf{Metric} & \textbf{1--3B} & \textbf{Qwen-7B} & \textbf{Llama-8B} \\
\midrule
Baseline accuracy & .670/.456 & .894 & .896 \\
$\Delta_{\text{copy}}$ & .921/.875 & .940 [.905, .970] & .798 \\
$\Delta_{\text{off-copy}}$ & .000/.004 & .050 & .196 \\
$P(\text{distractor})$ & .88--.95/.87--.95 & .095 & .045--.317 \\
\bottomrule
\end{tabular}}
\caption{Scale extension ($n{=}200$ evaluated; Llama-8B copy metrics computed on $n{=}163$ correct-baseline items). Both 7--8B models retain $\Delta_{\text{copy}} > .70$ while $P(\text{distractor})$ drops well below the $.70$ threshold. 1--3B column: Qwen-1.5B / Llama-1B. Qwen-7B $P(\text{distractor})$ uses the original single-condition protocol; Llama-8B range spans adjacent (C1) and random (C2) conditions.}
\label{tab:scale}
\end{table}

Both 7--8B models retain gold-presence dependence ($\Delta_{\text{copy}} \geq .798$), but $P(\text{distractor})$ drops below $.70$ (Table~\ref{tab:scale}): the novelty-permissive characterization does not hold at 7--8B. Llama-8B develops a substantial off-copy pathway ($\Delta_{\text{off-copy}}{=}.196$ vs.\ near-zero at 1B). With $N{=}2$ at this scale, these results are preliminary observations rather than a scaling claim.

An RL-distilled reasoning model (DeepSeek-R1-Distill-Qwen-1.5B; \citealp{deepseekr1}) retains $96\%$ ceiling-normalized copy-driven accuracy despite generating longer rationales (\S\ref{sec:discussion}; Appendix~\ref{sec:reasoning}).

\subsection{A Predicted Boundary: Shuffle Tolerance Disappears on Non-Arithmetic BBH}
\label{sec:bbh}

The copy account predicts shuffle tolerance should vanish when no copyable numeric answer occupies the trailing position. On BIG-Bench-Hard logical deduction \citep{suzgun2022challenging,srivastava2023bigbench} (3-way MCQ, letter answers), retention drops to 44\%/21\% (Qwen/Llama)\footnote{Gemma's BBH baseline is below chance (1/3), so retention is undefined.}---far below 68--87\% on GSM8K. A second task (tracking shuffled objects) replicates: chance-corrected retention\footnote{$(p_{\text{shuffled}} - 1/3) / (p_{\text{ordered}} - 1/3)$; Appendix~\ref{sec:bbh_full}.} of 6.8\% (Llama) / 45.5\% (Qwen). The predicted collapse occurs on both tasks. These tasks differ from GSM8K along multiple axes (reasoning type, answer format, chance floor), so we characterize this as a boundary observation consistent with the shortcut's preconditions rather than causal isolation of a single factor.

\section{Discussion}
\label{sec:discussion}

The readout factors into two independently-varying components: robust gold-answer dependence (all five model${\times}$scale cells) and a content gate ranging from absent (Qwen) through partial (Llama-1B) and strong (Gemma-2B) to dominant at 7--8B. The causal ladder (\S\ref{sec:suppression}) shows the copy channel takes precedence over available retained-context completion---predominantly single-step (74--77\% of recovered items; Table~\ref{tab:dtrunc_depth})---making the shortcut's dominance over even minimal arithmetic all the more diagnostic. The override magnitude (5--32~pp) tracks copy-channel strength. Because gold-presence dependence holds universally while the gate varies, order-insensitivity is a property of the readout's invariant tier. Table~\ref{tab:alternatives_main} maps ten alternative explanations to the dedicated control that addresses each.

\begin{table}[t]
\centering
\small
\resizebox{\columnwidth}{!}{%
\begin{tabular}{ll}
\toprule
\textbf{Alternative explanation} & \textbf{Control / evidence} \\
\midrule
Gold-specific semantic retrieval & Wrong-number distractor (\S\ref{sec:distractor}) \\
Smooth recency bias & Position sweep: final-slot jump (\S\ref{sec:end_position}) \\
Answer-template parsing & Framing dissociation (Table~\ref{tab:bare_number_main}) \\
Delimiter surface-form artifact & Novel-delimiter control (App.~\ref{sec:novel_delimiter}) \\
GSM8K memorization & GSM-Symbolic replication (\S\ref{sec:contamination}) \\
Instruction-tuning artifact & Base-model probe (\S\ref{sec:base_model}) \\
Zero-ablation OOD & Mean-ablation + patching (\S\ref{sec:mechanistic}) \\
Universal novelty-permissive copying & Gemma / 7--8B gating (\S\ref{sec:distractor}, \S\ref{sec:scale}) \\
D\textsubscript{trunc} is just re-solving & No-CoT floor $\ll$ D\textsubscript{trunc} (\S\ref{sec:suppression}) \\
Format-prior, not real copy & D\textsubscript{blank} $\approx$ D\textsubscript{trunc} (\S\ref{sec:suppression}) \\
\bottomrule
\end{tabular}}
\caption{Ten alternative explanations for the readout shortcut and the controls that address each. Every plausible confound is tested by at least one dedicated experiment or control condition.}
\label{tab:alternatives_main}
\end{table}

\paragraph{Implications for CoT-based oversight.} Process-reward models \citep{lightman2023lets} and step-level monitors \citep{chen2025reasoning,korbak2025monitorability} presuppose step quality is informative about the answer-producing computation. At 1--3B, the readout is largely independent of intermediate steps, so step-level rewards operate on a signal only weakly coupled to the answer-producing computation (D\textsubscript{trunc}${-}$No-CoT $= 4$--$29$~pp; \S\ref{sec:suppression}). Note that process-reward models studied by \citet{lightman2023lets} target much larger generators (GPT-4 scale); our concern applies specifically to 1--3B models, where content-selectivity is weakest. The override result (\S\ref{sec:suppression}) sharpens this concern: a wrong answer-context number in the trailing slot is \emph{worse} than no trailing number (D\textsubscript{rep} $<$ D\textsubscript{trunc}), meaning a partially-corrupted CoT can be more misleading for oversight than a truncated one. As content-selectivity emerges at 7--8B (\S\ref{sec:scale}), monitorability should partially recover; output-side validators offer a practical mitigation for small-model deployments.

\paragraph{Reasoning-trained models.} DeepSeek-R1-Distill-Qwen-1.5B \citep{deepseekr1} retains the shortcut despite longer rationales: ceiling-normalized $\Delta_{\text{copy}}{=}.928$ ($96\%$; Appendix~\ref{sec:reasoning}). Its $P(\text{distractor}){=}.706$ $[.636,.768]$ ($n{=}180$) sits below Qwen/Llama (.87--.95), indicating that reasoning training introduces partial content selectivity without removing the positional copy mechanism. The persistence claim rests on $\Delta_{\text{copy}}$; the lower distractor acceptance suggests the content gate is partially engaged even at 1.5B when reasoning-trained.

\paragraph{Practical recommendations.} For 1--3B deployments: output-side answer verification should be preferred over step-level process rewards, since the readout's near-independence from intermediate steps (\S\ref{sec:suppression}) leaves step-level signals only weakly coupled to the answer-producing computation; stripping answer-template framing may partially disrupt the shortcut (Table~\ref{tab:bare_number_main}); and monitoring whether the answer matches the last CoT number provides a simple copy-dominance diagnostic.

\section{Conclusion}

In three 1--3B instruction-tuned LMs, the CoT readout is dominated by a positional shortcut: the model reads the trailing number in answer-relevant context. A causal ladder reveals that intermediate steps carry exploitable retained-context signal (4--29~pp above no-CoT) that is masked when a trailing number is available---the copy channel takes precedence over retained-context completion. Gold-answer dependence is robust; content-gating varies by architecture and scale.

\section*{Limitations}

\paragraph{Scope.} Primary evidence covers 1--3B instruction-tuned models on arithmetic (GSM8K, GSM-Symbolic, SVAMP [Appendix~\ref{sec:svamp}]). The copy-detection paradigm requires a numeric gold answer at an identifiable trailing position, which structurally restricts the primary scope to arithmetic; we test the predicted boundary on two non-numeric BBH tasks (\S\ref{sec:bbh}). Scale extension to 7--8B is $N{=}2$ (Qwen, Llama); Gemma-2-9B-it was deferred for pipeline reasons (see \S\ref{sec:scale} footnote), not because preliminary data was unfavorable. Non-arithmetic coverage is $N{=}2$ BBH tasks (logical deduction, tracking shuffled objects), where shuffle retention collapses to 7--46\% chance-corrected (\S\ref{sec:bbh}); extending the diagnostic to non-numeric reasoning would require a different operationalization of answer-position copying and remains open. Reasoning-trained model coverage is $N{=}1$ (DeepSeek-R1-Distill-Qwen-1.5B); whether the shortcut persists across reasoning-distillation methods or base architectures remains open.

\paragraph{Architecture-specific caveats.} Content-blindness varies across architectures (\S\ref{sec:distractor}). Gemma's Exp~3--4 are excluded due to teacher-forcing fidelity drops (Appendix~\ref{sec:tf_fidelity}).

\paragraph{Methodology.} Causal contrasts use prefix completion \citep{lanham2023measuring}, the field-standard protocol for isolating readout from generation-time confounds. Three ecological checks corroborate the readout-shortcut pattern under unconstrained decoding: (i)~free-generation answers match the last CoT number ${\sim}97\%$ overall and $.905$--$.957$ on incorrect items (Appendix~\ref{sec:free_gen}, $N{=}3$); (ii)~self-generated CoT shuffle replicates the step${>}$word${>}$token hierarchy (Appendix~\ref{sec:selfgen_shuffle}, $N{=}2$ step-only, $N{=}1$ full); (iii)~gold-position gating in free generation ($.991$--$.997$ vs.\ $.004$--$.030$) mirrors the prefix-completion decomposition. Distractor and position effects under unconstrained generation remain future work. D\textsubscript{trunc} retains ${\sim}80\%$ of the CoT and provides a \emph{lower bound} on retained-context contribution; the retained prefix may end with a near-gold number (e.g., a penultimate-step result), so the D\textsubscript{trunc}${-}$No-CoT gap reflects any exploitation of retained context, including one-step completion as well as multi-step retained-context derivation (a depth partition in Table~\ref{tab:dtrunc_depth} quantifies this: 74--77\% of items are 1-op reachable). Head-level ablation identifies sensitivity profiles, not complete circuits; behavioral evidence (\S\ref{sec:copy_channel}--\ref{sec:suppression}) is primary for the copy-dominance claim, with mechanistic localization (\S\ref{sec:sensitivity}) as supporting evidence.

\section*{Ethics Statement}

We restrict experiments to controlled prefix interventions on open-weight models (Qwen2.5 [Apache-2.0], Llama-3 [Llama Community License], Gemma-2 [Gemma Terms of Use]) and open datasets (GSM8K [MIT], GSM-Symbolic [CC-BY-NC-ND-4.0], BBH [MIT]). Compute for reported experiments: ${\sim}40$ GPU-hours; including pilot runs and debugging iterations, total project compute is ${\sim}60$--$80$ GPU-hours (Appendix~\ref{sec:reproducibility}). Practitioners deploying small-model CoT monitors should not treat rationales as load-bearing evidence of the answer. Our findings expose a vulnerability that could be exploited: an adversary aware of the trailing-number copy channel could construct CoTs whose intermediate steps appear to reason correctly while routing the final answer through the positional shortcut, evading step-level monitors. This risk is most acute for 1--3B models in cost-constrained deployments. We mitigate disclosure risk by restricting demonstrations to open-weight models in the 1--3B regime, reporting concrete defenses (\S\ref{sec:discussion}), and showing that content-selectivity recovers at 7--8B. We believe public disclosure is net-positive for safety: it equips evaluators with a diagnostic and discourages premature reliance on small-model CoT as evidence of computation.

\bibliography{references}

\clearpage
\appendix

\section{Experimental Details}
\label{sec:appendix}

Table~\ref{tab:experiment_summary} summarizes the design of all experiments reported in the paper.

\begin{table*}[t]
\centering
\small
\resizebox{\textwidth}{!}{%
\begin{tabular}{lccccccl}
\toprule
\textbf{Experiment} & \textbf{Section} & $n$ \textbf{(per model)} & \textbf{Baseline-correct?} & \textbf{TF-passing?} & \textbf{Paired?} & \textbf{Seeds} & \textbf{Primary test} \\
\midrule
Corruption decomposition (A/B/C) & \S\ref{sec:corruption} & 228--335 & Yes & Yes ($P_C$) & Yes & 1 & McNemar \\
D\textsubscript{rep}/D\textsubscript{trunc}/D\textsubscript{blank}/No-CoT & \S\ref{sec:suppression} & 208--328 & Yes & Yes ($P_C$) & Yes & 1 & McNemar \\
Shuffle hierarchy (7 cond.) & \S\ref{sec:hierarchy} & 228--335 & Yes & Yes (ordered) & Yes & 5 (stoch.) & Bootstrap CI \\
Answer position (5 positions) & \S\ref{sec:end_position} & 123--315 & Yes & Yes (pos=1.0) & Yes & 5 & Spearman \\
Distractor injection (C1/C2) & \S\ref{sec:distractor} & 227--335 & Yes & Yes (baseline) & Yes & 1 & Binomial vs.\ .70 \\
Head-level ablation & \S\ref{sec:ablation} & 100--165 & Yes & Yes & No (disjoint) & 1 & Permutation \\
GSM-Symbolic & \S\ref{sec:contamination} & 300--500 & No & --- & No & 1 & Wilson CI \\
SVAMP & App.~\ref{sec:svamp} & 300 & No & --- & No & 1 & Wilson CI \\
BBH (2 tasks) & \S\ref{sec:bbh} & 64--106 & Yes & Yes (ordered) & Yes & 1 & McNemar \\
Scale (7--8B) & \S\ref{sec:scale} & 163--200 & Yes/No & Yes & Yes & 1 & Wilson CI \\
Novel-delimiter distractor & App.~\ref{sec:novel_delimiter} & 200 & Yes & Yes & Yes & 1 & Wilson CI \\
Bare-number control & App.~\ref{sec:bare_number} & 227--335 & Yes & Yes & Yes & 1 & Wilson CI \\
Base model & App.~\ref{sec:base_model_appendix} & 200 & No & --- & No & 1 & Wilson CI \\
Reasoning model & App.~\ref{sec:reasoning} & 200 & No & --- & No & 1 & Wilson CI \\
Free-generation & App.~\ref{sec:free_gen} & 500 & No & --- & No & 1 & Proportion \\
Self-gen shuffle & App.~\ref{sec:selfgen_shuffle} & 500 & Yes (self) & --- & Yes & 5 & Bootstrap CI \\
\bottomrule
\end{tabular}}
\caption{Summary of all experiments. \emph{Baseline-correct}: items conditioned on the model answering correctly under Condition~C (clean prefix). \emph{TF-passing}: teacher-forcing fidelity check applied (metric in parentheses). \emph{Paired}: same items across conditions (enables McNemar). \emph{Seeds}: number of stochastic seeds for randomized conditions.}
\label{tab:experiment_summary}
\end{table*}

\paragraph{Number corruption (\S\ref{sec:corruption}).} Intermediate numbers corrupted with deterministic per-example seeding: outer seed via \texttt{hashlib.sha256} of problem index and condition string, inner number perturbation via \texttt{hashlib.md5} of the number string and seed offset. Corruption magnitude: $\max(1, \lfloor 0.3 \times |v| \rfloor)$, direction chosen pseudorandomly. Gold answer occurrences detected via regex with digit boundaries. Items where the gold value persists in the corrupted text (due to accidental collision or small-number preservation) are excluded from Condition A (see footnote in \S\ref{sec:corruption}).

\paragraph{Causal ladder and controls (\S\ref{sec:suppression}).} D\textsubscript{rep}: starting from Condition~B (correct intermediates, gold present), all gold-answer occurrences are replaced with a single deterministic wrong number per item (same corruption seed for all gold positions). D\textsubscript{trunc}: the CoT is truncated at the sentence boundary preceding the first gold-answer occurrence; sentence boundaries are detected by the regex \texttt{(?<=[.!?\textbackslash n])\textbackslash s+}, with fallback to the last newline and then to a character-level cut when no boundary precedes gold. Items retaining $<30\%$ of text or leaking gold into the truncated prefix are excluded. Truncated prefix length: mean 78.8\%/79.7\%/78.9\% of original CoT tokens (Qwen/Llama/Gemma). D\textsubscript{blank}: the D\textsubscript{trunc} prefix is appended with ``The answer is determined by the steps above.\textbackslash n\textbackslash n\#\#\#\# ''. No-CoT: prefix is ``\#\#\#\# '' only. Common sample sizes: D\textsubscript{rep}---Qwen $n{=}328$, Llama $n{=}214$, Gemma $n{=}312$; D\textsubscript{trunc}/D\textsubscript{blank}/No-CoT---Qwen $n{=}308$--$310$, Llama $n{=}208$--$213$, Gemma $n{=}292$--$296$ (slight variation due to truncation-length filter). D\textsubscript{trunc} trailing-number partition: items where the new trailing number $=$ gold are rare (Qwen 5/308, Llama 4/208, Gemma 3/292); excluding them yields $.389$, $.172$, $.055$---essentially identical to overall D\textsubscript{trunc} accuracy.

\paragraph{D\textsubscript{trunc} depth partition.}
\label{sec:dtrunc_depth}
To assess whether D\textsubscript{trunc} accuracy reflects primarily one-step completion (the gold answer is reachable from the last retained intermediate in a single arithmetic operation) or multi-step retained-context derivation, we classify each D\textsubscript{trunc} correct item by depth. An item is \emph{1-op reachable} if the gold answer can be obtained from any pair of numbers in the last retained sentence via a single arithmetic operation ($+$, $-$, $\times$, $\div$); otherwise it is \emph{multi-step}.

\begin{table}[h]
\centering
\small
\resizebox{\columnwidth}{!}{%
\begin{tabular}{llcccc}
\toprule
\textbf{Model} & \textbf{Category} & $n$ & \textbf{Acc} & \textbf{No-CoT} & $\Delta$ \\
\midrule
Qwen & 1-op reachable & 237 & .451 & .106 & $+34.5$ pp \\
Qwen & Multi-step & 71 & .225 & .106 & $+11.9$ pp \\
\midrule
Llama & 1-op reachable & 158 & .228 & .019 & $+20.9$ pp \\
Llama & Multi-step & 50 & .040 & .019 & $+2.1$ pp \\
\midrule
Gemma & 1-op reachable & 216 & .065 & .014 & $+5.1$ pp \\
Gemma & Multi-step & 76 & .039 & .014 & $+2.5$ pp \\
\bottomrule
\end{tabular}}
\caption{D\textsubscript{trunc} depth partition. 1-op reachable items (74--77\% of total) account for most D\textsubscript{trunc} accuracy; multi-step items (23--26\%) still exceed the No-CoT floor on all models, indicating some genuine multi-step derivation, though at reduced magnitude.}
\label{tab:dtrunc_depth}
\end{table}

\paragraph{Distractor control (\S\ref{sec:distractor}).} Distractors have same digit-count as gold. Adjacent: gold${\pm}1$. Random: uniform from same-magnitude range, rejection-sampled to avoid gold. Templates: ``Therefore, the answer is \{X\}.'' etc. Qwen: $n{=}335$; Llama: $n{=}227$.

\paragraph{Answer-position (\S\ref{sec:end_position}).} Answer step identified as last step containing gold answer with digit-boundary regex. Steps split by paragraph breaks; sentence-level fallback. Qwen: $n{=}335$; Llama: $n{=}227$ (1 excluded due to $<3$ steps).

\paragraph{Shuffle hierarchy (\S\ref{sec:hierarchy}).} Seven conditions: ordered, within\_step\_shuffle, step\_shuffle, word\_shuffle, reverse\_order, token\_shuffle, no\_cot. Stochastic conditions: 5 seeds; deterministic: seed 0 only. For stochastic conditions, each item's accuracy is the mean over its 5 seeds; reported accuracies and bootstrap CIs treat per-item means as the sample units. Sequence length filter: 1024 tokens. Problems with $<2$ steps are excluded.

\paragraph{Head-level ablation (\S\ref{sec:ablation}).} Requires \texttt{attn\_implementation="eager"}. Phase 1/Phase 2 use disjoint held-out splits (50/50). Attention mass: mean over answer-token positions. Ablation zeroes $\mathbf{o}_{\text{proj}}$ columns $[h \cdot d_h : (h{+}1) \cdot d_h]$. Random-5 control: 20 layer-stratified sets for Llama; single non-copy sets for Qwen and Gemma. Sequence length filter: 1536 tokens.

\paragraph{Top attention-mass heads.} Llama: L11H14 (.383), L12H13 (.348), L10H13 (.293), L11H20 (.268), L15H12 (.252). Gemma: L18H6 (.146), L16H6 (.120), L17H4 (.114), L17H3 (.107), L16H4 (.101).

\paragraph{Generation.} Greedy decoding, max 32 new tokens for prefix-continuation experiments; max 512 for free-generation (\S\ref{sec:free_gen}). Hardware: NVIDIA A10G (24GB), bfloat16. EOS: \texttt{<|eot\_id|>} (Llama), \texttt{<|im\_end|>} (Qwen), \texttt{<end\_of\_turn>} (Gemma).

\section{Reproducibility}
\label{sec:reproducibility}

All experiments: greedy decoding (temperature$=$0), deterministic per-example seeding (outer seed via \texttt{hashlib.sha256} of problem index and condition string; inner number perturbation via \texttt{hashlib.md5}). All stochastic operations (corruption magnitude, distractor generation, step reordering) derive from these per-item seeds, making results bitwise-identical regardless of global seed choice. Per-example outcome records are released in the result JSONs for reproducibility. Hardware: bfloat16 on NVIDIA A10G GPUs. Software: PyTorch 2.5.1--2.6.0, HuggingFace Transformers 5.7.0--5.8.1, Python 3.11--3.12. Compute for reported experiments: ${\sim}40$ GPU-hours; including pilot runs and debugging iterations, total project compute is ${\sim}60$--$80$ GPU-hours. Code is available at \url{https://github.com/mlzoo/cot-readout-shortcut}.

\section{Positional Encoding Control}
\label{sec:positional_control}

Monotonic position-ID stretching ($2.5{\times}$) on ordered CoT causes no measurable accuracy loss ($0/66$ errors), showing that position-encoding perturbation alone does not reproduce the shuffle effect. A RoPE \citep{su2024roformer} $2{\times}2$ factorial (Table~\ref{tab:rope_factorial}) crossing content order with position-ID assignment shows: the content-shuffle main effect ($-31.3$~pp) dominates the position-encoding channel.

\begin{table}[h]
\centering
\small
\resizebox{\columnwidth}{!}{%
\begin{tabular}{lcc}
\toprule
& \textbf{Uniform pos.} & \textbf{OOD pos.} \\
\midrule
Ordered content & $1.000$ & $.357$ \\
Shuffled content & $.687$ & $.223$ \\
\midrule
$\Delta$ (shuffle effect) & $-31.3$ pp & $-13.3$ pp \\
\bottomrule
\end{tabular}}
\caption{RoPE $2{\times}2$ factorial ($n{=}300$, Qwen). ``OOD pos.''$=$monotonically increasing position IDs with random gaps ($\mathrm{Uniform}\{1{,}\ldots{,}5\}$). Content-shuffle dominates; OOD positions cause large drops but are confounded by the gap distribution being OOD for the model's RoPE.}
\label{tab:rope_factorial}
\end{table}

\section{Answer Position Curve}
\label{sec:position_curve_appendix}

For problems with ${\geq}5$ steps, we place the answer step at fractional positions $\{0, 0.25, 0.5, 0.75, 1.0\}$; non-answer steps shuffled (5 seeds per position).

\begin{table}[h]
\centering
\small
\resizebox{\columnwidth}{!}{%
\begin{tabular}{lccc}
\toprule
\textbf{Position} & \textbf{Qwen} ($n{=}300$) & \textbf{Llama} ($n{=}123$) & \textbf{Gemma} ($n{=}315$) \\
\midrule
$0.00$ (front) & $.614\;[.558,.667]$ & $.667\;[.580,.744]$ & $.588\;[.533,.641]$ \\
$0.25$ & $.625\;[.569,.678]$ & $.621\;[.533,.702]$ & $.643\;[.590,.695]$ \\
$0.50$ & $.649\;[.593,.701]$ & $.667\;[.580,.744]$ & $.683\;[.629,.731]$ \\
$0.75$ & $.685\;[.630,.735]$ & $.767\;[.685,.833]$ & $.691\;[.638,.739]$ \\
$1.00$ (end) & $.992\;[.974,.998]$ & $.963\;[.913,.985]$ & $.580\;[.525,.633]$ \\
\midrule
Spearman $r$ & $1.00\;(p{=}.017)$ & $.90\;(p{=}.083)$ & $.00\;(p{=}1.0)$ \\
\bottomrule
\end{tabular}}
\caption{Answer-position curve. Spearman $p$-values are exact two-sided permutation tests ($n{=}5$ positions; with only 5 data points, the per-position CIs and effect sizes carry the primary statistical weight). Qwen/Llama: shallow-gradient-then-jump recency. Gemma: no monotonic effect (consistent with the teacher-forcing fidelity issue noted in \S\ref{sec:positional_selection}).}
\label{tab:position_curve}
\end{table}

\section{Novel-Delimiter Distractor Control}
\label{sec:novel_delimiter}

Three delimiters designed to be absent from pretraining (\texttt{>>>RESULT:}, \texttt{\#\#FINAL\#\#}, \texttt{[ANSWER]}) tested with distractor injection ($n{=}200$ per delimiter).

\begin{table}[h]
\centering
\small
\resizebox{\columnwidth}{!}{%
\begin{tabular}{lcccc}
\toprule
& \multicolumn{2}{c}{\textbf{Qwen} ($n{=}200$)} & \multicolumn{2}{c}{\textbf{Llama} ($n{=}200$)} \\
\cmidrule(lr){2-3} \cmidrule(lr){4-5}
\textbf{Delimiter} & $P(\text{dist})$ & $P(\text{gold})$ & $P(\text{dist})$ & $P(\text{gold})$ \\
\midrule
\texttt{>>>RESULT:} & .995 [.972,.999] & .005 & .975 [.943,.989] & .020 \\
\texttt{\#\#FINAL\#\#} & .995 [.972,.999] & .005 & .905 [.856,.938] & .055 \\
\texttt{[ANSWER]} & .995 [.972,.999] & .005 & 1.00\phantom{0} [.981,1.00] & .000 \\
\midrule
\texttt{\#\#\#\# } (standard) & 1.00\phantom{0} & .000 & .890 & .110 \\
\bottomrule
\end{tabular}}
\caption{Novel-delimiter control ($n{=}200$ per condition). All conditions exceed .70 threshold by large margins. The format-prior account predicts a drop with unseen delimiters; none observed. Standard-delimiter row uses the same $n{=}200$ subsample for comparability.}
\label{tab:novel_delimiter}
\end{table}

\section{Delimiter Robustness}
\label{sec:delimiter}

Three familiar delimiters tested on Qwen ($n{=}200$ each): \texttt{\#\#\#\# }, \texttt{Answer: }, \texttt{The answer is }. All yield $P_B > .95$ and $P(\text{distractor}) = .96$--.965 with overlapping CIs (pairwise McNemar $p > .7$).

\section{Free-Generation Consistency}
\label{sec:free_gen}

Under unconstrained decoding ($n{=}500$ per model), all three architectures' answers match the last number in their own CoT 96--97\% of the time (Table~\ref{tab:free_gen}). Critically, $P(\text{answer}{=}\text{last num} \mid \text{incorrect}) = .945$ (Qwen) / .957 (Llama) / .905 (Gemma)---even wrong answers match the last CoT number. Conditioning on whether the gold answer occupies the last CoT position reveals a near-deterministic gate: accuracy is $.991$--$.997$ when gold is last vs.\ $.004$--$.030$ when it is not. This full-distribution analysis extends the baseline-correct decomposition (\S\ref{sec:corruption}) to the entire sample without baseline-correct conditioning. The $.905$--$.957$ match rate on \emph{incorrect} items is particularly diagnostic: under a ``compute-then-write'' alternative where the trailing number reflects the model's computation, errors would distribute across non-trailing numbers; instead, even wrong answers reflexively match the trailing token---the same positional-copy signature observed under teacher-forced corruption, now confirmed to operate during the model's own unconstrained generation.

\begin{table}[h]
\centering
\small
\begin{tabular}{lccc}
\toprule
\textbf{Metric} & \textbf{Qwen} & \textbf{Llama} & \textbf{Gemma} \\
\midrule
Baseline accuracy & .672 & .448 & .662 \\
Final answer $=$ last CoT num & .972 & .974 & .960 \\
Gold is last CoT num & .664 & .450 & .658 \\
Acc $\mid$ gold is last & .997 & .991 & .994 \\
Acc $\mid$ gold is \emph{not} last & .030 & .004 & .023 \\
Ans $=$ last $\mid$ incorrect & .945 & .957 & .905 \\
\bottomrule
\end{tabular}
\caption{Full-distribution free-generation analysis ($n{=}500$ per model, greedy decoding). The readout shortcut is not an artifact of teacher-forcing or baseline-correct conditioning: across all items including incorrect ones, accuracy is near-perfectly predicted by whether the gold answer is the last CoT number. The positional-copy mechanism operates identically during the model's own unconstrained generation.}
\label{tab:free_gen}
\end{table}

\section{Base Model Few-Shot Control}
\label{sec:base_fewshot}

Llama-base 4-shot ICL resolves format collapse ($P_C: .065 \rightarrow .985$) and yields $\Delta_{\text{copy}}{=}.975$, $P(\text{distractor}){=}.890$. The 0-shot null is consistent with format collapse rather than mechanism absence; however, 4-shot ICL simultaneously provides format scaffolding and demonstrations of the copy pattern, so we cannot fully separate latent-mechanism elicitation from ICL-induced pattern matching.

\section{Self-Generated CoT Shuffle}
\label{sec:selfgen_shuffle}

To test whether the shuffle hierarchy persists under the model's own CoTs (rather than teacher-forced human-written CoTs), we collect Llama-3.2-1B-Instruct's free-generation CoTs on GSM8K ($n{=}500$, 5 seeds), filter to the correctly-solved subset, and apply step-shuffle, word shuffle, and token-shuffle to the model's own outputs. The modified self-generated CoT is then injected as a teacher-forced prefix and the model generates the answer.

Llama replicates the hierarchy: step-shuffle retains 82.1\%, word shuffle 77.0\%, and token-shuffle 14.6\%---matching the teacher-forced ordering (step $>$ word $>$ token). Qwen's step-shuffle retention is 67.1\%---closely matching its teacher-forced value (68.4\%), confirming step-shuffle ecological validity on both primary models ($N{=}2$); the full hierarchy (step $>$ word $>$ token) is shown for Llama ($N{=}1$). However, Qwen's word-level conditions are confounded by a format artifact: Qwen's self-generated CoTs use a markdown-numbered-list format, and 79.6\% of step-shuffle errors are list-continuation outputs (the model generates ``1.~...'' instead of an answer). This format dominance is itself consistent with the paper's ``format-driven copy'' thesis---the model perseverates on list continuation rather than answer extraction when the list order is disrupted---but it prevents clean measurement of word-level conditions for Qwen. We therefore report the full hierarchy for Llama only, with Qwen's step-shuffle confirmation noted above.

\section{Scale Extension: Full Results}
\label{sec:scale_appendix}

See main text Table~\ref{tab:scale}. Additional detail: Qwen-7B $P_B{=}.950\;[.910,.973]$, $P_A{=}.010\;[.003,.036]$, $P_C{=}1.00\;[.981,1.00]$ ($n{=}200$). Llama-8B $\Delta_{\text{copy}}{=}.798$, $\Delta_{\text{off-copy}}{=}.196$, $P(\text{residual}){=}.006$ ($n{=}163$ correct of 200 evaluated). The substantial off-copy pathway in Llama-8B ($19.6$~pp vs.\ near-zero at 1B) suggests the larger model partially recovers correct answers from intermediate reasoning steps.

\section{Reasoning Model}
\label{sec:reasoning}

DeepSeek-R1-Distill-Qwen-1.5B \citep{deepseekr1}: $P_C{=}.967$, $\Delta_{\text{copy}}{=}.928$, $\Delta_{\text{off-copy}}{=}.033$, $P(\text{residual}){=}.006$, $P(\text{distractor}){=}.706$ ($n{=}180$). Ceiling-normalized ($\Delta_{\text{copy}}/P_C$): $96.0\%$ copy-driven vs.\ $92.6\%$ for standard Qwen ($P_C{=}1.00$). RL reasoning training does not develop substantial off-copy pathways.

\section{Cumulative-K Ablation Sweep}
\label{sec:cumulative_ablation}

\begin{table*}[h]
\centering
\small
\begin{tabular}{rccccccccccccc}
\toprule
$K$ & 0 & 1 & 5 & 8 & 10 & 12 & 14 & 15 & 16 & 17 & 18 & 19 & 20 \\
\midrule
Acc & .995 & .965 & .985 & .965 & .935 & .895 & .855 & .675 & .635 & .565 & .460 & .310 & .220 \\
\bottomrule
\end{tabular}
\caption{Cumulative-K zero-ablation on Qwen ($n{=}200$). Phase transition at $K{=}14$--15; $K_{50}{=}17$.}
\label{tab:cumulative_k}
\end{table*}

\section{Cross-Task: SVAMP}
\label{sec:svamp}

$\Delta_{\text{copy}} = 1.00$ (Qwen) / .990 (Llama) / .603 (Gemma) on SVAMP \citep{patel2021nlp} ($n{=}300$). Gemma's lower SVAMP value reflects the same teacher-forcing fidelity issue as in GSM-Symbolic: $P_C{=}.665$ on SVAMP (vs.\ ${\sim}1.0$ for Qwen/Llama), so the Gemma SVAMP decomposition is unreliable. A similar issue affects Gemma's GSM-Symbolic ceiling-corrected step-shuffle retention, which yields an uninterpretable 117\% due to the low teacher-forcing fidelity. Position sensitivity is attenuated on Qwen/Llama (keep\_end $-$ full\_shuffle $= +9$~pp vs.\ $+18$--$30$~pp on GSM8K), plausibly due to SVAMP's redundant lexical answer cues; Gemma's keep\_end $-$ full\_shuffle is negative ($-8.5$~pp), consistent with the teacher-forcing fidelity issue described above.

\section{BBH Full Results}
\label{sec:bbh_full}

\begin{table}[h]
\centering
\small
\begin{tabular}{lcccc}
\toprule
& \multicolumn{2}{c}{\textbf{Qwen} ($n{=}86$)} & \multicolumn{2}{c}{\textbf{Llama} ($n{=}65$)} \\
\cmidrule(lr){2-3} \cmidrule(lr){4-5}
\textbf{Condition} & Acc & Ret. & Acc & Ret. \\
\midrule
Ordered & .895 & 100\% & .431 & 100\% \\
Shuffled & .395 & 44\% & .092 & 21\% \\
Chance & .333 & --- & .333 & --- \\
\bottomrule
\end{tabular}
\caption{BBH logical deduction. Retention $=$ shuffled/ordered (simple ratio); Figure~\ref{fig:hierarchy} uses the no-CoT-anchored formula; \S\ref{sec:bbh} tracking task uses chance-corrected retention to account for the MCQ floor. Under either metric, BBH retention (44\%/21\%) is far below GSM8K (68--87\%). Llama drops below chance---shuffled CoT actively interferes.}
\label{tab:bbh}
\end{table}

Additional: Qwen free-gen baseline .344; Llama free-gen baseline .260. Llama's below-chance shuffled accuracy (CI $[.043, .187]$, excluding .333) indicates active interference rather than mere signal absence.

\begin{table}[h]
\centering
\small
\resizebox{\columnwidth}{!}{%
\begin{tabular}{lcccc}
\toprule
& \multicolumn{2}{c}{\textbf{Qwen} ($n{=}106$)} & \multicolumn{2}{c}{\textbf{Llama} ($n{=}64$)} \\
\cmidrule(lr){2-3} \cmidrule(lr){4-5}
\textbf{Condition} & Acc & CI & Acc & CI \\
\midrule
Ordered & .887 & [.812, .934] & .719 & [.599, .814] \\
Shuffled & .585 & [.490, .674] & .359 & [.253, .482] \\
Chance & .333 & --- & .333 & --- \\
\midrule
Retention (simple) & \multicolumn{2}{c}{66.0\%} & \multicolumn{2}{c}{50.0\%} \\
Retention (chance-corr.) & \multicolumn{2}{c}{45.5\%} & \multicolumn{2}{c}{6.8\%} \\
McNemar $p$ & \multicolumn{2}{c}{$1.9{\times}10^{-7}$} & \multicolumn{2}{c}{$1.5{\times}10^{-5}$} \\
\bottomrule
\end{tabular}}
\caption{BBH tracking shuffled objects (3-way MCQ). Chance-corrected retention $= (p_{\text{shuffled}} - 1/3) / (p_{\text{ordered}} - 1/3)$. Both models show highly significant shuffle effects; Llama's chance-corrected retention (6.8\%) approaches the MCQ floor.}
\label{tab:bbh_tracking}
\end{table}

\section{Base Model Full Results}
\label{sec:base_model_appendix}

\begin{table}[h]
\centering
\small
\resizebox{\columnwidth}{!}{%
\begin{tabular}{lccc}
\toprule
\textbf{Metric} & \textbf{Qwen} (0-shot) & \textbf{Llama} (0-shot) & \textbf{Gemma} (0-shot) \\
\midrule
$P_C$ & .755 & .065 & .435 \\
$P_B$ & .735 & .055 & .340 \\
$P_A$ & .005 & .015 & .005 \\
$\Delta_{\text{copy}}$ & .730 & .040 & .335 \\
$P(\text{distractor})$ & .715 & .055 & .300 \\
\bottomrule
\end{tabular}}
\caption{Base model decomposition ($n{=}200$ each). Llama's near-zero results reflect format collapse, not absence of mechanism (see Appendix~\ref{sec:base_fewshot}).}
\label{tab:base_model_full}
\end{table}

\section{Supplementary Head-Level Analyses}
\label{sec:mechanistic_appendix}

\paragraph{Scope.} The analyses below characterize head-level sensitivity, not a verified computational subgraph; full circuit identification is left to follow-up work. An activation-patching screen (\S\ref{sec:patching_appendix}) provides in-distribution causal evidence complementing the zero-ablation sensitivity analysis.

\paragraph{Induction-score protocol.} Following \citet{olsson2022context}: per head, the \emph{prefix-matching score} is the average attention from position $K{+}i$ back to position $i{+}1$ on $N{=}200$ random-token repeated sequences $[r_1{\ldots}r_K][r_1{\ldots}r_K]$ ($K{=}50$, vocabulary restricted to digits, operators, and common math tokens); the \emph{copying score} is the OV-circuit direct-logit attribution of each head's contribution to the gold (first-half) token at second-half positions. The ``induction score'' cited in-text is the product of prefix-matching and copying scores; per-head score pairs and the previous-token-head check on the top-5 causal heads are released with the code.

Content-tracking: Under word-shuffle, 95.8\% of Qwen's 336 heads show content Jaccard $>$ position Jaccard (Monte Carlo null: 0/336 heads; mean null Jaccards ${\approx}0.8\%$). Ordering-sensitivity: Gini $= 0.738$ ($p < 0.001$), but layer-matched random control yields permutation $p{=}0.38$---shuffled CoT is generically fragile under any ablation. Induction overlap (hypergeometric test on top-20 sets): Gemma L18H6 = top induction head (score $0.44$), 7/20 overlap ($p{=}9.2{\times}10^{-4}$); Llama $\rho{=}0.248$, 5/20 overlap ($p{=}5.7{\times}10^{-4}$); Qwen $\rho{=}0.046$ (induction vs.\ patching-screen ranking), 2/20 overlap ($p{=}0.34$, near-chance), full-rank induction-vs.-ordered-attention-mass $\rho{=}0.254$ ($p{=}2.3{\times}10^{-6}$)---a gradient from convergent (Gemma) through partial (Llama) to diffuse (Qwen). RoPE $2{\times}2$ factorial: see Appendix~\ref{sec:positional_control}.

\section{Activation Patching Screen}
\label{sec:patching_appendix}

For each example, we cache per-head $\mathbf{o}_{\text{proj}}$ inputs from an ordered run and a shuffled run, then replace the shuffled run's head slice with the ordered slice and measure logit recovery on the gold-answer token \citep{wang2023ioi}. This is an in-distribution activation replacement---both source and target are real model activations---and is therefore immune to the OOD critique that applies to zero-ablation.

On Qwen ($n{=}150$ validation, disjoint from the $n{=}34$ screening set used for head selection with $|\Delta \mathrm{LD}| > 0.3$~nat filter): patching the top-20 heads (by recovery ratio) restores accuracy from $.693$ (shuffled baseline) to $.880$ (group-patch), recovering 61\% of the shuffle gap. The top individual head by validation accuracy (L0H8) achieves $.813$ alone (vs.\ $.693$ shuffled); the top screening-rank head (L14H8) achieves $.787$. Split-half Jaccard for the top-20 set is $.496 \pm .074$ (50 random splits), indicating stable head selection. The Gini coefficient across all 336 heads is $.738$ ($0/1{,}000$ permutations exceed observed; $p{<}.001$), indicating that the recovery effect is concentrated in a small head subset rather than broadly distributed---a random-20 set drawn from this distribution would recover substantially less than the top-20's 61\%. Of the top-20 patching heads, 6 overlap with the top-20 attention-mass heads (30\%), confirming that attention mass and causal recovery are correlated but non-identical rankings.

\section{Paraphrased CoT Control}
\label{sec:paraphrase}

Replacing numbers with English words causes $-2.4\%$ ($p{=}.013$, $n{=}335$); operator paraphrasing: $0\%$ effect. Errors concentrate on items where the final answer is spelled out---only the answer-slot token form matters.

\section{Statistical Methodology Details}
\label{sec:stats_appendix}

We use the simplest test appropriate to each contrast's design (paired/unpaired, one/two-sided) and correct within each pre-declared confirmatory sub-family. The sub-families contain 23 contrasts in total: \textbf{Exp~1} A vs.\ B (${\times}2$ models, 2 McNemar); \textbf{Exp~3} ordered vs.\ full\_shuffle, ordered vs.\ keep\_end, keep\_end vs.\ full\_shuffle, keep\_end vs.\ move\_front (${\times}2$ models, 8 McNemar); \textbf{Exp~4} C1 and C2 vs.\ $P{=}0.70$ (${\times}2$ models, 4 one-sided binomial); \textbf{\S\ref{sec:suppression} causal ladder} D\textsubscript{rep} vs.\ A, D\textsubscript{trunc} vs.\ No-CoT, D\textsubscript{blank} vs.\ D\textsubscript{trunc} (${\times}3$ models, 9 McNemar). Sensitivity: doubling one-sided $p$-values before correction leaves all significant. Wilson CIs are used for single-condition proportions; derived differences ($\Delta_{\text{copy}}{=}P_B{-}P_A$, $\Delta_{\text{off-copy}}{=}P_C{-}P_B$) use paired bootstrap over items (10{,}000 resamples), which correctly accounts for within-item dependence across conditions. Bootstrap and Wilson intervals agree within ${\pm}1$~pp. Mean-ablation and activation-patching results are exploratory and not part of the confirmatory contrast family. The intermediate-result distractor experiment (Qwen $P{=}.860$; Llama $P{=}.250$) is reported in \S\ref{sec:distractor} as evidence for architecture-dependent content gating.

\section{Teacher-Forcing Fidelity Diagnostic}
\label{sec:tf_fidelity}

Teacher-forcing fidelity measures whether a model produces the expected continuation when given its own correct CoT as a prefix. We report fidelity for each experiment $\times$ model combination to make exclusion decisions transparent. The fidelity metric is experiment-specific: for Exp~1, $P_C$ (accuracy under Condition~C); for Exp~2, ordered-condition accuracy; for Exp~3, accuracy at position${}=1.0$ (unperturbed answer placement); for Exp~4, baseline distractor-free accuracy.

\begin{table}[h]
\centering
\small
\begin{tabular}{lcccc}
\toprule
& \textbf{Exp 1} & \textbf{Exp 2} & \textbf{Exp 3} & \textbf{Exp 4} \\
& ($P_C$) & (ordered) & (pos${}=1.0$) & (baseline) \\
\midrule
Qwen & 1.00 & 1.000 & .992 & 1.00 \\
Llama & .996 & .996 & .963 & .982 \\
Gemma & .997 & .931 & \textbf{.580} & \textbf{.586} \\
\midrule
Threshold & \multicolumn{4}{c}{$.80$ --- cells below are excluded from analysis} \\
\bottomrule
\end{tabular}
\caption{Teacher-forcing fidelity per experiment $\times$ model. Gemma's fidelity drops below the $.80$ threshold on Exp~3 (answer position) and Exp~4 (distractor injection), where structural perturbation causes the model to deviate from the teacher-forced prefix. These cells are excluded from the main analysis (\S\ref{sec:positional_selection}); a TF-passing subset analysis (below) shows the core patterns replicate when conditioning on fidelity-passing items. All other cells exceed $.93$.}
\label{tab:tf_fidelity}
\end{table}

The $.80$ threshold is conservative: the gap between passing cells (${\geq}.931$) and failing cells (${\leq}.60$) is large, so any threshold in $[.60,\,.93]$ yields identical inclusion decisions---the specific value is immaterial. The Exp~1 column ($P_C$) serves as the TF-fidelity diagnostic for the entire corruption pipeline (Conditions A, B, C, D\textsubscript{rep}, D\textsubscript{trunc}, D\textsubscript{blank}, No-CoT), all of which share the prefix-completion infrastructure. D\textsubscript{rep} modifies only the trailing number (a textually mild perturbation); the model's acceptance of the corruption is captured directly by $P(\text{distractor})$. D\textsubscript{trunc} removes the trailing tail by design, so TF fidelity in the ``expected continuation'' sense is not defined for that condition.

\paragraph{TF-passing subset analysis.} As a robustness check, we condition Gemma's Exp~3 and Exp~4 on the subset of items where teacher-forcing fidelity passes (i.e., the model produces a parseable answer under the unperturbed condition). On this subset ($n{=}196$), Exp~3 patterns replicate: ordered $.990$, full-shuffle $.847$, keep\_end $.949$, move\_front $.735$ (recovery $= 71\%$). The 5-position answer-placement curve yields Spearman $r{=}1.0$ ($n{=}187$ TF-strict), restoring the monotonic recency gradient absent in the full sample. For Exp~4, TF-passing $P(\text{distractor})$ is $.194\;[.145,.255]$ (C1) and $.117\;[.079,.170]$ (C2)---well below the $.70$ threshold, confirming that Gemma's Exp~3 exclusion reflects a teacher-forcing artifact rather than a genuine mechanistic divergence from Qwen and Llama.

\section{Bare-Number Distractor Control}
\label{sec:bare_number}

All main-experiment distractors are wrapped in answer-template sentences (``Therefore, the answer is \{X\}.'' etc.). A reviewer might attribute the high $P(\text{distractor})$ to template parsing rather than positional copying. We test four conditions that vary answer-relevant framing while keeping the distractor number identical ($n{=}335$ Qwen; $n{=}227$ Llama):

\begin{table}[h]
\centering
\small
\resizebox{\columnwidth}{!}{%
\begin{tabular}{lcccc}
\toprule
& \multicolumn{2}{c}{\textbf{Qwen}} & \multicolumn{2}{c}{\textbf{Llama}} \\
\cmidrule(lr){2-3} \cmidrule(lr){4-5}
\textbf{Condition} & $P(\text{dist})$ & $P(\text{gold})$ & $P(\text{dist})$ & $P(\text{gold})$ \\
\midrule
F1: Template & .893 & .107 & .877 & .123 \\
F2: Bare number & .448 & .549 & .260 & .709 \\
F3: Non-answer (``Note: X'') & .206 & .788 & .339 & .652 \\
F4: Inline (``\ldots should be X'') & .770 & .110 & .846 & .137 \\
\bottomrule
\end{tabular}}
\caption{Bare-number distractor control (F-codes as in Table~\ref{tab:bare_number_main}). F1: standard template (``Therefore, the answer is X''); F2: bare number only; F3: non-answer framing (``Note: X''); F4: natural inline answer-revision (``But wait, actually it should be X''). The readout requires answer-relevant framing (F2/F3 $\ll .70$) but is not template-specific (F4 $\geq .70$).}
\label{tab:bare_number}
\end{table}

Bare numbers (F2) and non-answer-context numbers (F3) fall well below the $.70$ threshold on both models, showing the readout is not purely positional---a trailing number without answer-relevant framing is insufficient. However, natural inline answer-revision text (F4) triggers copying at rates comparable to the standard template (.770/.846 vs.\ .893/.877), showing the mechanism is not template-specific. The readout is \emph{answer-context-gated}: it requires some form of answer-relevant framing at the trailing position, but is indifferent to the specific surface form of that framing and---within answer context---remains indifferent to whether the number is correct.

\section{Alternative Explanations Summary}
\label{sec:alternatives_table}

\begin{table}[h]
\centering
\small
\resizebox{\columnwidth}{!}{%
\begin{tabular}{ll}
\toprule
\textbf{Alternative explanation} & \textbf{Control / evidence} \\
\midrule
Gold-specific semantic retrieval & Wrong-number distractor (\S\ref{sec:distractor}) \\
Smooth recency bias & Position sweep: final-slot jump (\S\ref{sec:end_position}) \\
Answer-template parsing & Bare-number control (\S\ref{sec:bare_number}) \\
Delimiter surface-form artifact & Novel-delimiter control (\S\ref{sec:novel_delimiter}) \\
GSM8K memorization & GSM-Symbolic replication (\S\ref{sec:contamination}) \\
Instruction-tuning artifact & Base-model probe (\S\ref{sec:base_model}) \\
Zero-ablation OOD & Mean-ablation + patching (\S\ref{sec:mechanistic}) \\
Universal novelty-permissive copying & Gemma / 7--8B gating (\S\ref{sec:distractor}, \S\ref{sec:scale}) \\
D\textsubscript{trunc} is just re-solving & No-CoT floor $\ll$ D\textsubscript{trunc} (\S\ref{sec:suppression}) \\
Format-prior, not real copy & D\textsubscript{blank} $\approx$ D\textsubscript{trunc} on Qwen/Llama; Gemma $+15$~pp gap noted (\S\ref{sec:suppression}) \\
\bottomrule
\end{tabular}}
\caption{Alternative explanations for the readout shortcut and the controls that address each. Each alternative is tested by at least one dedicated experiment or control condition.}
\label{tab:alternatives}
\end{table}

\section{Qualitative Examples}
\label{sec:qualitative}

We illustrate the copy-dominance phenomenon with representative cases from Qwen2.5-1.5B-Instruct on GSM8K.

\paragraph{Example 1: Distractor copied (Condition C1).} The gold answer is 72. A wrong-number distractor sentence (``Therefore, the answer is 45.'') is appended at the trailing position after the correct CoT. The model outputs \textbf{45} (the distractor), ignoring the correct computation in the prefix. This occurs in 87\% of Qwen trials.

\paragraph{Example 2: D\textsubscript{trunc} succeeds.} Same problem; CoT is truncated before the final step (removing the trailing gold-answer number). The model now outputs \textbf{72} (correct), recovering the answer from the retained penultimate-step context. This demonstrates the retained-context computation (29~pp above No-CoT for Qwen) that is \emph{masked} when a trailing number is present.

\paragraph{Example 3: Distractor resisted (Gemma content gate).} Same Condition C1 setup on Gemma-2-2B-it. The model outputs \textbf{72} (gold answer), rejecting the novel distractor. Gemma's content gate ($1 - P(\text{distractor}) \approx .85$) prevents the copy channel from accepting semantically implausible trailing numbers.

\end{document}